\documentclass[journal,comsoc]{IEEEtran}

\usepackage[T1]{fontenc}

\usepackage{ifpdf}
\usepackage{cite}
\usepackage{graphicx}
\graphicspath{{./Fig/}}
\usepackage{amsmath,amsfonts}
\usepackage{bm}
\usepackage{algorithm}
\usepackage{algorithmic}
\usepackage{array}
\usepackage[caption=false,font=normalsize,labelfont=sf,textfont=sf]{subfig}
\usepackage{url}
\begin{document}

\title{Characterization and Design of A Hollow Cylindrical Ultrasonic Motor}

\author{Zhanyue Zhao, Yang Wang, Charles Bales, Daniel Ruiz-Cadalso, Howard Zheng, Cosme Furlong-Vazquez, Gregory Fischer
\thanks{Zhanyue Zhao, Yang Wang, Charles Bales, and Gregory Fischer are with the Department of Robotics Engineering, Worcester Polytechnic Institute, Worcester, MA 01605 USA (e-mail: zzhao4@wpi.edu, gfischer@wpi.edu).}
\thanks{Daniel Ruiz-Cadalso, Howard Zheng, and Cosme Furlong-Vazquez are with the Department of Mechanical \& Materials Engineering, Worcester Polytechnic Institute, Worcester, MA 01605 USA.}
\thanks{This research is supported by National Institute of Health (NIH) under the National Cancer Institute (NCI) under Grant R01CA166379 and R01EB030539.}
}


\maketitle

\begin{abstract}

Piezoelectric ultrasonic motors perform the advantages of compact design, faster reaction time, and simpler setup compared to other motion units such as pneumatic and hydraulic motors, especially its non-ferromagnetic property makes it a perfect match in MRI-compatible robotics systems compared to traditional DC motors. Hollow shaft motors address the advantages of being lightweight and comparable to solid shafts of the same diameter, low rotational inertia, high tolerance to rotational imbalance due to low weight, and tolerance to high temperature due to low specific mass. This article presents a prototype of a hollow cylindrical ultrasonic motor (HCM) to perform direct drive, eliminate mechanical non-linearity, and reduce the size and complexity of the actuator or end effector assembly. Two equivalent HCMs are presented in this work, and under 50g prepressure on the rotor, it performed 383.3333rpm rotation speed and 57.3504mNm torque output when applying 282$V_{pp}$ driving voltage.

\end{abstract}


\section{Introduction}

\IEEEPARstart{R}{obot-assisted} procedures have played an important role in recent years, especially in the MRI environment considering its limited space for operation. Our previous work represents the MRI-compatible robot applied in various procedures, and in some MR-guided robot-assisted surgery that uses a needle for a specific procedure, it needs to be placed above the target tumor with delicate control \cite{nycz2017mechanical,gandomi2020modeling,campwala2021predicting,szewczyk2022happens,tavakkolmoghaddam2021neuroplan,jiang2024icap,tavakkolmoghaddam2023passive,tavakkolmoghaddam2023design,zhao2024deep,zhao2024development}, under this circumstance the size and design of a needle manipulator remain challenging. A hollow shaft motor is a type of motor with a tube shaft instead of a solid cylinder shaft. The hollow shaft motors are widely used in research and manufacturing areas, with the advantage of (a) lightweight and comparable to the solid shaft of the same diameter, (b) low rotational inertia, (c) cheaper, (d) high tolerance to rotational imbalance due to low weight, and (e) be tolerant to high temperature due to low specific mass. However, some disadvantages exist for hollow shafts, which are (a) less tolerant to over-torque, (b) prone to bending especially when exposed to lateral stress, and (c) even though it is tolerant to mild imbalance, this however increases as rotational speed exceeds the critical limit. The hollow shaft motor (HSM), or hollow cylindrical motor (HCM), is expected to improve control precision through the removal of backlash from any gearing. Allowing the actuator to go through the hollow shaft will produce further advantages, (a) directly driving without a belt or gearbox, (b) enables high speed, and excellent response time and eliminates mechanical nonlinearities introduced by gearboxes, and (c) reduces the size and complexity of actuator or end effector assembly. 

Several studies have explored various designs of hollow core piezoelectric motors and their performance capabilities. Lu \textit{et al.} developed an in-plane rotary ultrasonic motor that utilizes four separately placed Langevin-type bending vibrators around a ring-shaped stator. This design achieved a no-load speed of approximately 100 rpm and a stalling torque of 0.3 Nm, with a maximum rotating speed of 255 rpm at an input voltage of 300 V for a ball-shaped metal rotor\cite{lu2014novel}. Yang \textit{et al.} introduced a cylindrical traveling wave ultrasonic motor employing a bonded-type composite beam and a novel excitation mode for longitudinal-bending (L-B) hybrid vibrations. This configuration requires only two PZT ceramic plates and a single metal beam to generate synchronous L-B vibrations with a constant 90° phase difference when two alternating voltages with phase differences are applied, resulting in a maximum no-load speed of 342 rpm and an output torque of 6.26 mNm at 100 V\cite{yang2016cylindrical}.

In the realm of combined rotary-linear piezoelectric actuation, Mashimo \textit{et al.} designed a piezoelectric actuator capable of both rotational and translational movements using a single stator. Operating at the frequency of the R3 mode and the common frequency of the T1 and T2 modes, this actuator achieved a maximum rotational speed of approximately 160 rpm and a linear speed of about 63 mm/s at an applied voltage of 42 V\cite{mashimo2009rotary}. Tuncdemir \textit{et al.} developed a multi-degree-of-freedom (MDoF) ultrasonic motor with a cylindrical-joint configuration that combines rotary and translational motions. The prototype, measuring 5 mm in diameter and 25 mm in length, attained translational speeds of 5 mm/s and rotational speeds of 3 rad/s under a 4 mN blocking force when driven by a 20 V square wave input signal\cite{tuncdemir2011design}. Additionally, Elbannan \textit{et al.} designed a two-degree-of-freedom inchworm actuator based on earlier concepts\cite{salisbury2006design}, producing a prototype compatible with MRI environments that delivers linear motion speeds of 5.4 mm/s and rotational speeds of 10.5 rpm\cite{elbannan2012design, el2015development}.

In this work, we extended our previous work in \cite{zhao2023preliminary} by adding tapered rotors frictionally coupled with two equivalent hollow cylindrical ultrasonic motors (HCM) and validating with speed and torque measurements. Under the parameter of 50g prepressure on the rotor, the HCM performed 383.3333rpm rotation speed and 57.3504mNm torque with 282$V_{pp}$ driving voltage.

\section{Hollow Cylindrical Motor Design}

\begin{figure}
\centerline{\includegraphics[width=0.9\linewidth]{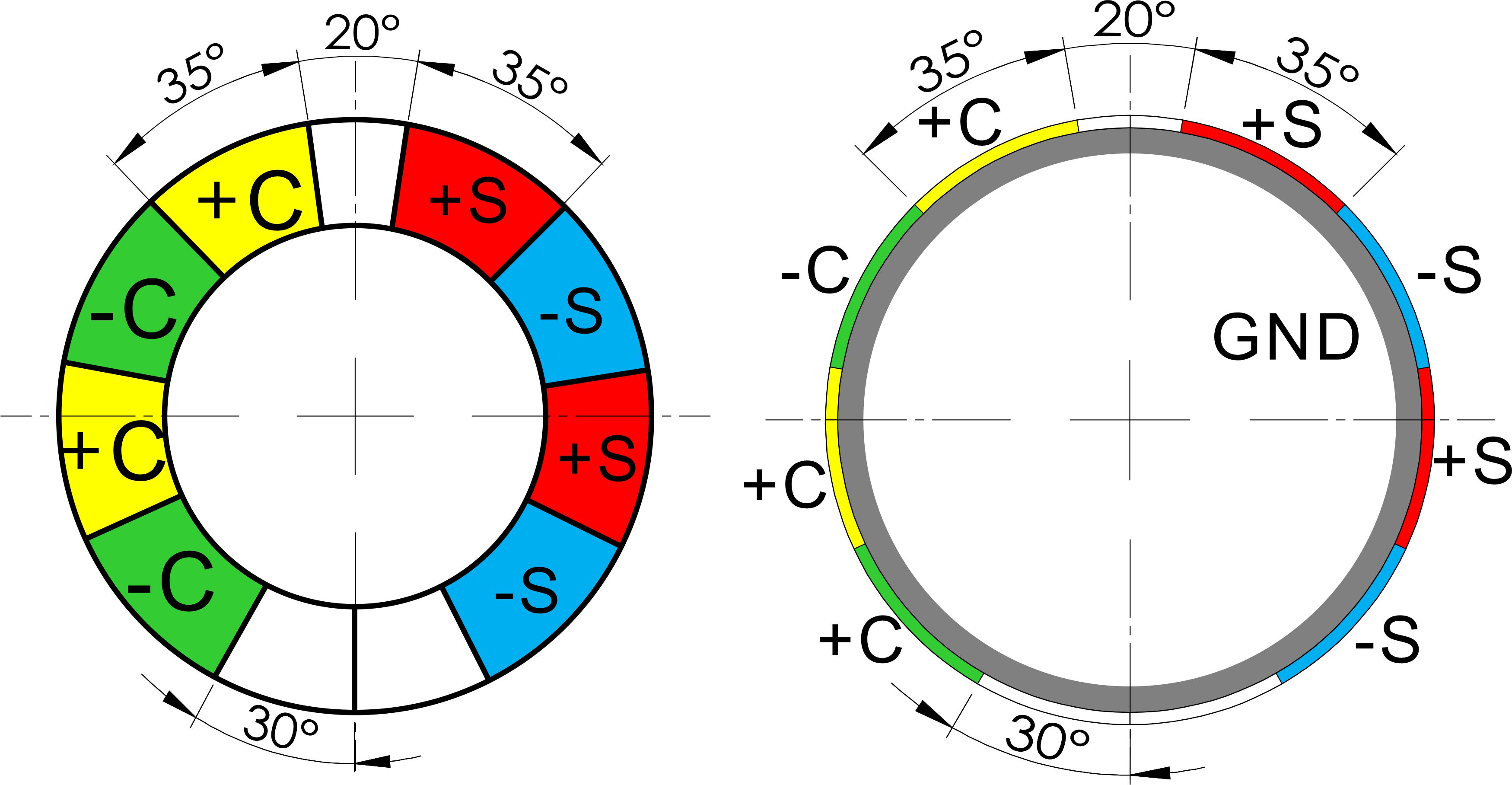}}
\caption{HCM PZT pattern with internal design. (Left) Standard 4-phase type motor flat ceramic ring etching pattern. (Right) Vertical flipped ceramic tube etching type for internal HCM design.}
\label{fig:hollowpattern}
\end{figure}

PZT-5H ceramic was used with the etching pattern shown in Fig \ref{fig:hollowpattern}. Instead of a flat ring, we used a vertical tube as the main driving component, which can be considered flipping the flat ring to a vertical setup. The shifted voltage signals and GND are applied to the vertical wall of the etched ceramic tube. Here is the internal design of the ceramic tube, the voltage signals are applied on the external lateral etched wall while the GND is applied on the whole internal wall. The voltage will apply through the wall and generate motion to and outwards the tube center line. Detailed description can be found in the previous study \cite{zhao2023preliminary}.

\begin{figure}
\centerline{\includegraphics[width=0.9\linewidth]{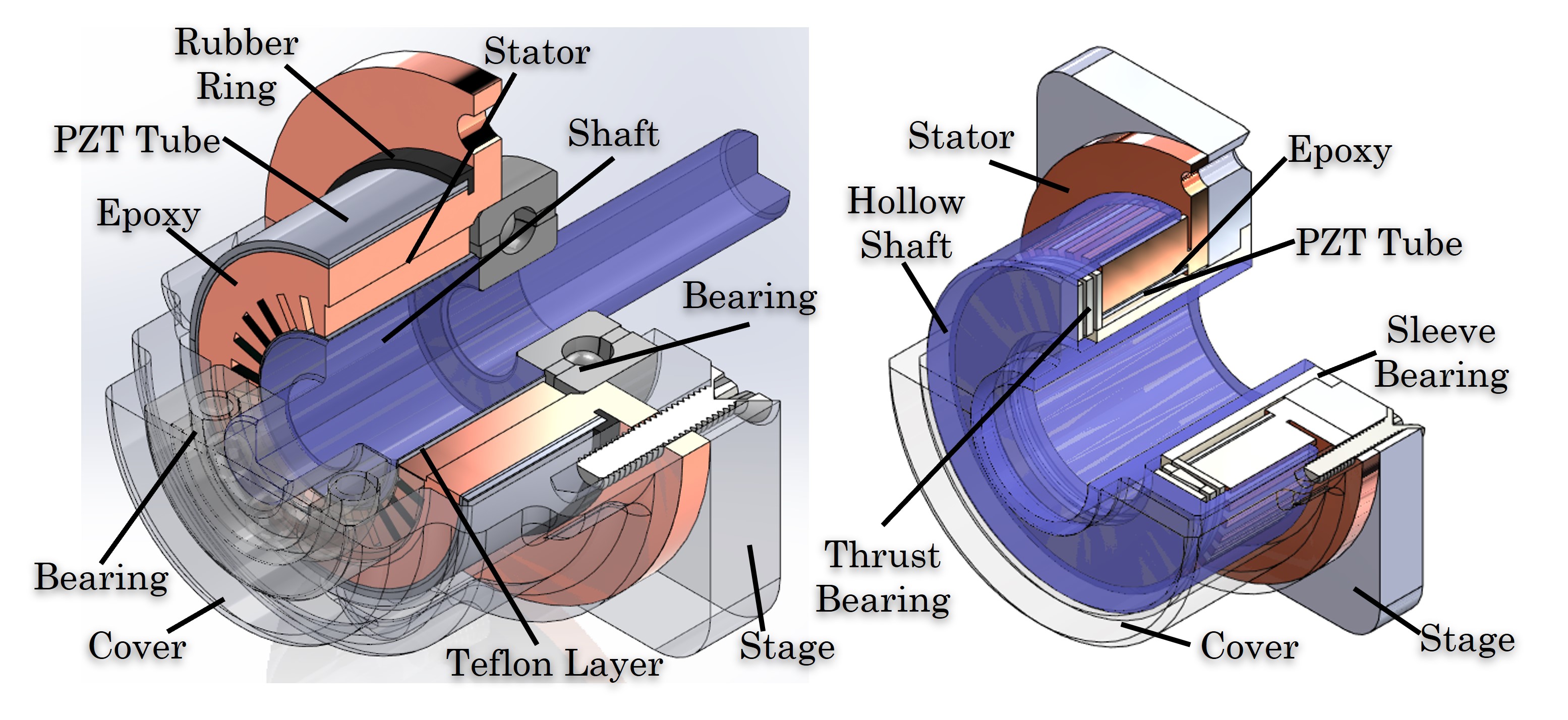}}
\caption{Two types of HCM design, (Left) Internal type design and  (Right) External design.}
\label{fig:design}
\end{figure}

Based on the tube etch pattern (internal or external), the HCM can be designed into two different types, internal and external design, which are shown in Fig \ref{fig:design}. The internal design on the left indicates the stator's teeth are towards the center of the ceramic tube, and frictionally coupled with the shaft acting as the rotor. To increase the friction between the stator and the rotor, a layer of Teflon film is applied to the rotor shaft engage section and is directly in contact with stator teeth. An etched PZT-5H tube with Fig \ref{fig:hollowpattern} etch pattern on the external wall is attached to the outer circumference of the stator with Loctite\textregistered~3888 electrically-conductive epoxy adhesive. Two rotary bearings hold the rotor shaft top and bottom for motion balance, and the stage and cover enclose the stator and bearings. The external design on the right indicates the stator's teeth are towards the outside of the ceramic tube, also frictionally coupled with a hollow rotor drum. An etched PZT-5H tube with an opposite etching pattern compared to Fig \ref{fig:hollowpattern} with an etching pattern on the internal wall is attached to the inner circumference of the stator with 3888 epoxy. A sleeve bearing and a thrust bearing are holding the hollow rotor shaft both axially and radially. A cover and stage enclose the motor with the rotor shaft output part outside for transmission.

Both designs depend on the traveling waves on the lateral wall of the stator transferring to the teeth to drive the rotor. To achieve the best performance and prove the concept, the internal design is used in this article, because of the below advantages, (a) easy to fabricate, especially since the stator can be made by wire Electrical discharge machining (EDM), (b) rotor is easy to manufacture, and (c) easy to apply pressure between stator and rotor.

\begin{figure}
    \centerline{\includegraphics[width=1\linewidth]{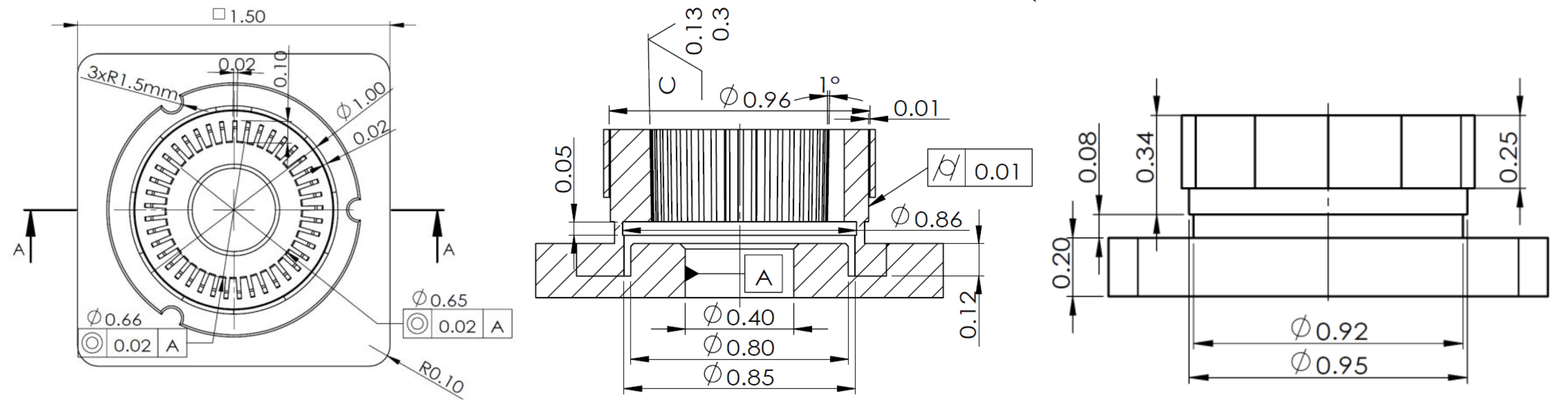}}
    \caption{Detailed geometry design of internal HCM.}
    \label{fig:geometry}
\end{figure}

The detailed design is shown in Fig \ref{fig:geometry}. A PZT-5H ceramic tube with 1.00 inch OD $\times$ 0.020-inch thick wall $\times$ 0.25-inch height, and electroless copper electrodes from EBL Products\textregistered~have etched the pattern shown in Fig \ref{fig:hollowpattern} on the external wall, a detailed etching method can be found in our previous work \cite{zhao2023preliminary,dadkhah2016self,dadkhah2018increasing}. The stator top is 0.95 inch OD with 0.01-inch tolerance for conductive epoxy, and the outer circumference requires 0.01-inch cylindricity. There are 22 $\times$ 0.02 $\times$ 0.10-inch notches with 1$^\circ$ chamfer (taper) from the top for rotor and stator contacting prepressure applying. The top edge of the teeth requires 0.02-inch concentricity based on the base center hole, while the teeth surface requires Ra 0.13 to 0.3-micron surface roughness. There is a 0.08 height $\times$ 0.03-inch wall thickness thin tube layer for less constrain from the base and better traveling waves transmission performance. The stator sits into the stage with 3 $\times$ R1.5mm half holes for the fixture. The stage size is a 1.50-inch edge square with a $\phi$0.40 inch center hole and a 0.20-inch height with a 0.12-inch depth notch for the stator to sit into.

\section{Stator Finite Element Modeling}

The finite element modeling method is used in the early stage of the stator preliminary study by Comsol Multiphysics 5.6. Two study types were discussed, which were eigenfrequency and time-dependent studies. The piezoelectric effect was used in the simulation, and then geometry and material were created based on the dimension and selected material from the library, namely copper (C360). In physics nodes, the solid mechanics and electrostatics were created based on the piezoelectric effect multiphysics node. In the solid mechanics node, the tube was added to the piezoelectric, and the bottom surface of the stator was defined as a fixed constraint. In the electrostatics node, GND and 4-channel phase-shifted voltage signals were applied onto the surface of the PZT tube, and the 4 voltage signals selected as electric potential are defined in Equation \ref{equ:signal}.  For the meshes section, we chose free tetrahedral with fine-level meshing for the whole body. The last step was defining the study and parameters, and the eigenfrequency and time dependent study were created. In the eigenfrequency study, we chose 30kHz as the target median frequency to look for other eigenfrequencies with other setups as default. After acquiring the specific eigenfrequencies, we chose the 5$^{th}$ frequency as the driving frequency used in time dependent study and set output times from 0 to 2.5 ms with 10 over the selected frequency, which is 10 times of period as incremental value. Detailed parameters can be found in previous work \cite{carvalho2020study,zhao2023preliminary,zhao2021preliminary,zhao2024study}.

\begin{equation}
    \begin{aligned}
        & V_{sin+} = V_0sin(2\pi f) \\
        & V_{cos+} = V_0sin(2\pi f+\pi/2)\\
        & V_{sin-} = -V_0sin(2\pi f)\\
        & V_{cos-} = -V_0sin(2\pi f+\pi/2)\\
    \end{aligned}
    \label{equ:signal}
\end{equation}

\begin{figure}
    \centerline{\includegraphics[width=0.9\linewidth]{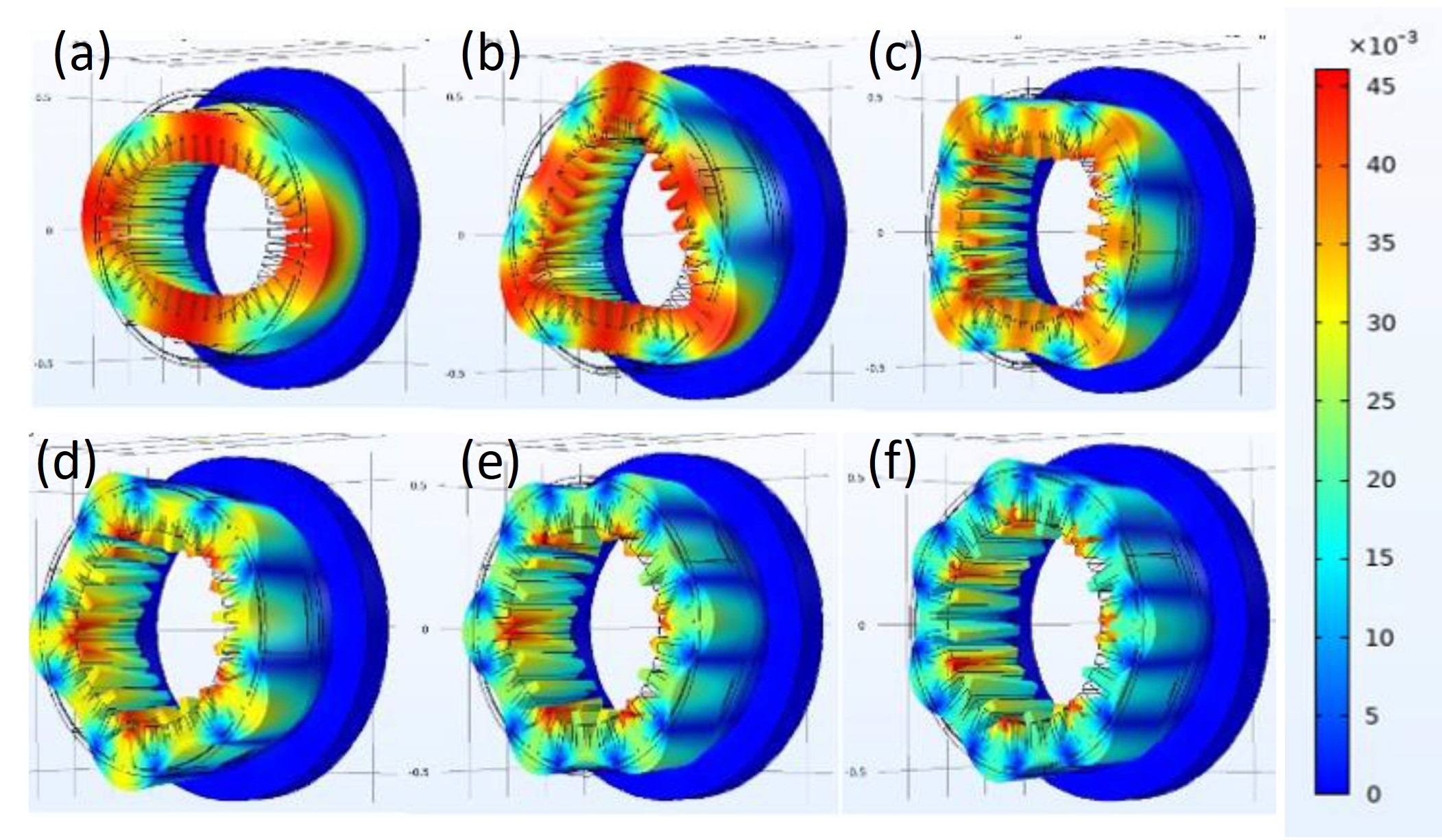}}
    \caption{(a)-(f) indicate copper stator from $2^{nd}$ to $7^{th}$ eigenfrequency simulation results. Legend unit is \emph{mm}.}
    \label{fig:coppersimu}
\end{figure}

Fig \ref{fig:coppersimu} shows the eigenfrequency for the stator with dimensions mentioned in \cite{zhao2023preliminary}. The simulation was not able to find the first excitation mode, and it shows the excitation pattern from the second mode. From (a) to (f) indicates copper stator from 2$^{nd}$ to 7$^{th}$ mode simulation results. From the pattern images, we can see the traveling waves were performing on the internal wall of teeth, and from top to bottom the wave amplitude is decreasing, the more close to the fixture stage the smaller wave amplitude will be generated. This is because the stage is fixed and constrained at the bottom which will reduce the motion close to the bottom section. Detailed value of eigenfrequency versus maximum towards center displacement can be found in Table \ref{tab:coppersimu}. 

The result shows that HCM towards the center distance is between 45.6770 to 46.3130 $\mu$m, which we ignore the mode 2 resonate frequency at 12.4812kHz performing 56.1362 $\mu$m. Under this frequency, there are only two contact areas and the rotor will not be kept stable operation. In this work, we choose mode 5 resonate frequency to operate both HCMs with output performance measurement.

\begin{table}[]
\centering
\caption{Stator uses C360 copper stimulation on excitation frequency and maximum towards center displacement.}
\label{tab:coppersimu}
\begin{tabular}{c|c|c}
\hline
Mode\# & Eigen Frequency (kHz) & Maximum Displacement ($\mu$m)\\ \hline
2 & 12.4812 & 56.1362 \\ \hline
3 & 21.8119 & 45.6770 \\ \hline
4 & 30.5002 & 45.9448 \\ \hline
5 & 39.8604 & 45.7191 \\ \hline
6 & 49.1277 & 46.3130 \\ \hline
7 & 63.8972 & 45.9928 \\ \hline
\end{tabular}
\end{table}

The time dependent study will compute a few important motor parameters including settling time and different section displacement on the teeth surface. The motor is considered settled when the sine amplitude stops increasing, and the motor settles to a steady state within around 0.6ms.

\section{Experiment Results and Analysis}
\label{exp}

\begin{figure}
    \centerline{\includegraphics[width=0.9\linewidth]{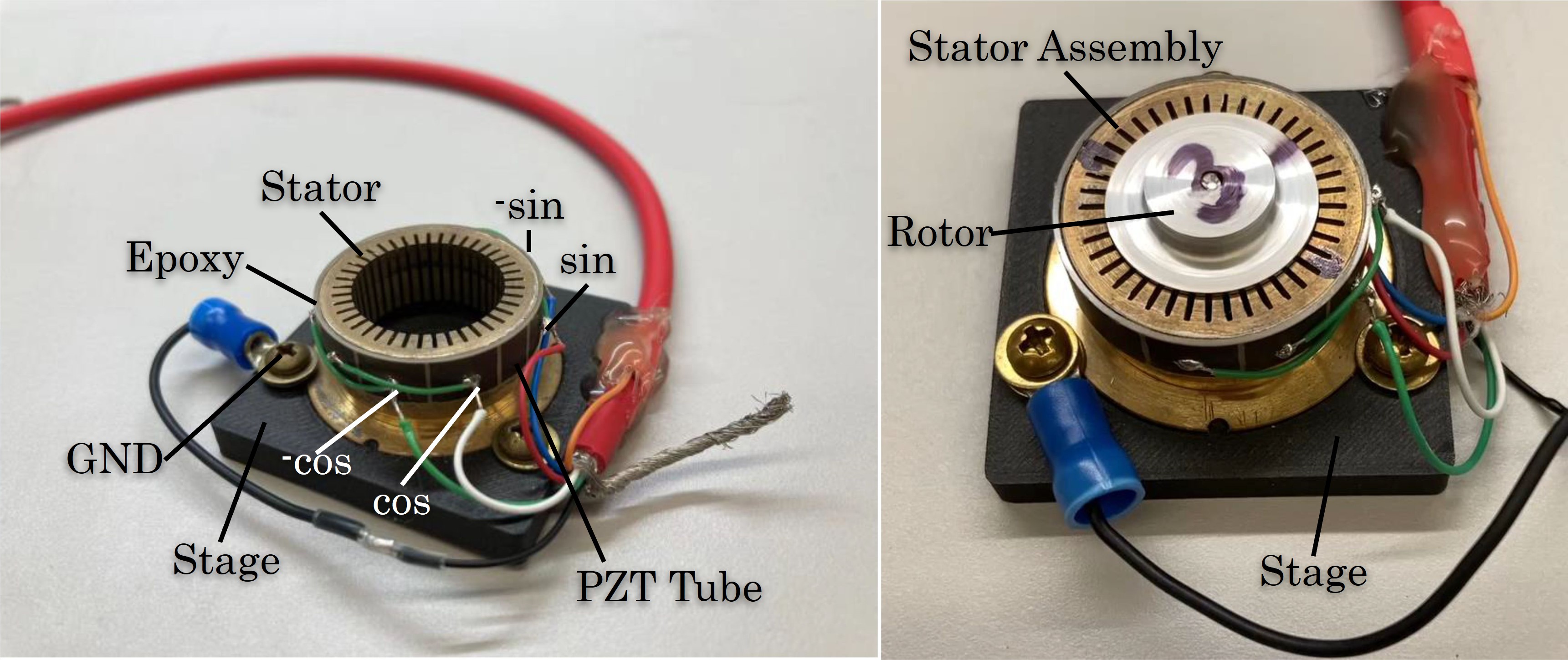}}
    \caption{(Left) One of the two equivalent HCMs stator assembly and cable connection 4-channel setup. (Right) An aluminum rotor coupled with the stator and rotary motion is performed.}
    \label{fig:hcmreal}
\end{figure}

Based on the simulation, two equivalent HCMs prototypes with C360 copper will be studied in this part. Fig \ref{fig:hcmreal} shows one of the two equivalent HCMs stator assemblies. We used C360 copper as the main stator material and used wire EDM technology to cut the inner teeth. An etched PZT-5H tube was attached outside the stator using conductive epoxy, where the whole stator is GND. 4 channels with 90$^\circ$ phase shift high voltage signals were soldered on the copper electrodes, and the GND terminal was connected physically by a screw on the side of the stator. The whole stator was seated into a carbon-fiber printed stage and fixed by 3 screws. An aluminum-made rotor with 1$^\circ$ opposite chamfer was fabricated to match the stator. 

\subsection{Holography Evaluation}

\begin{figure}
    \centerline{\includegraphics[width=0.9\linewidth]{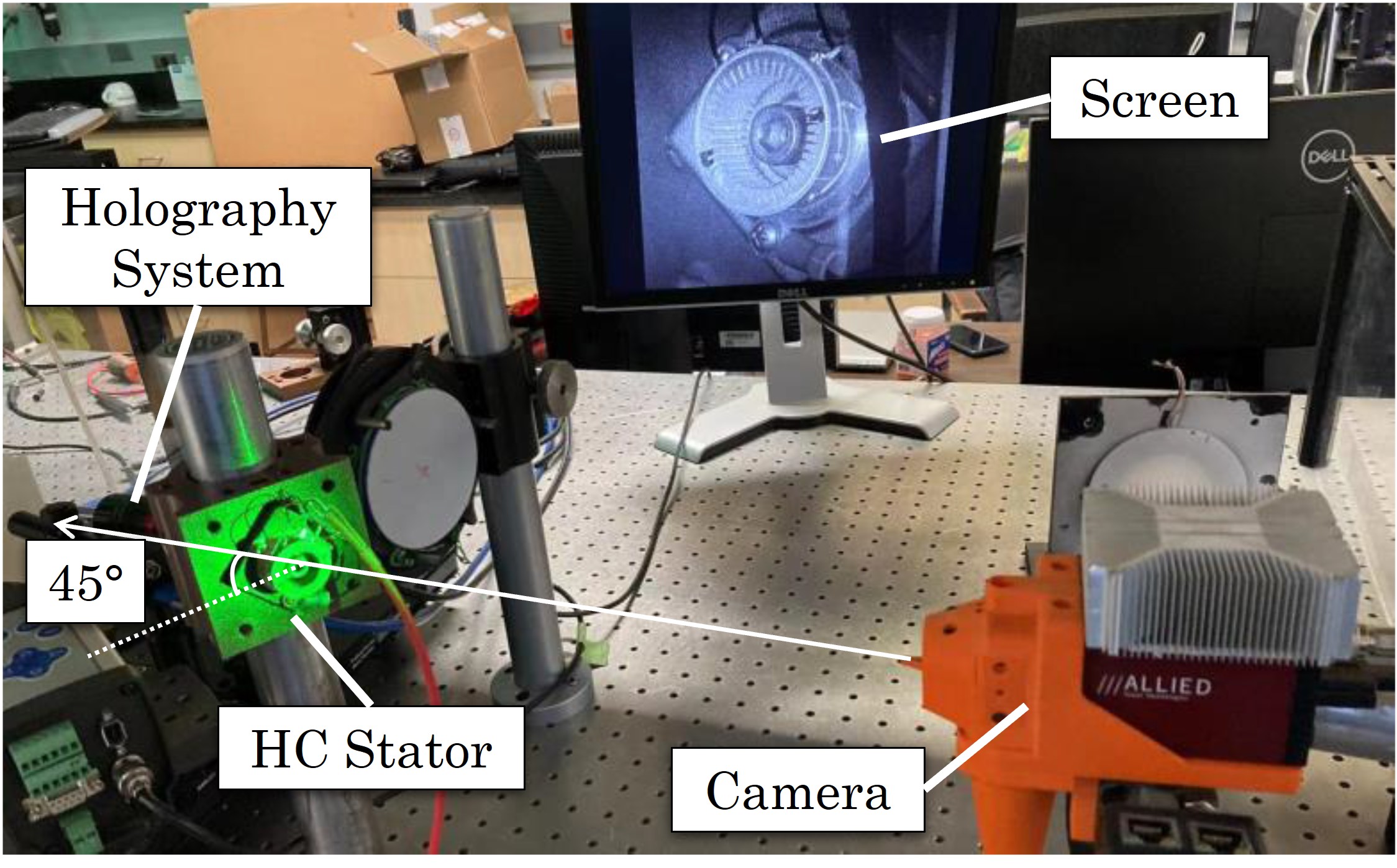}}
    \caption{The holography imaging system. The motor was fixed on the optical stage with $~45^\circ$ degrees along the camera direction so that the motion on the surface of the teeth could be observed.}
    \label{fig:holosetup}
\end{figure}

To evaluate the performance of HCMs, a digital holographic imaging system was used in time-averaged mode to capture the deformations induced in the stator by the excitation of the stator. The final setup is shown in Fig \ref{fig:holosetup}. The frequency of the different operating modes of the stator was determined by sweeping over the frequency range from 1 kHz to 60 kHz and recording the frequencies of the highest deformation. Considering the holography imaging method can observe a specific surface out-of-plane motion, this is not suitable for HCMs with toward center motion. By turning the stator by 45$^\circ$, the inner teeth surface acted as a sloping plane facing the camera, and the system can observe the motion. Notice that because the surface is not a flat surface that is perpendicular to the camera direction, the layer in the holography images is not very accurate, however, it is sufficient to visualize the excitation modes and standing wave formation. A detailed description can be found in \cite{zhao2023preliminary}.

\begin{figure*}
    \centering
    \includegraphics[width=0.8\textwidth]{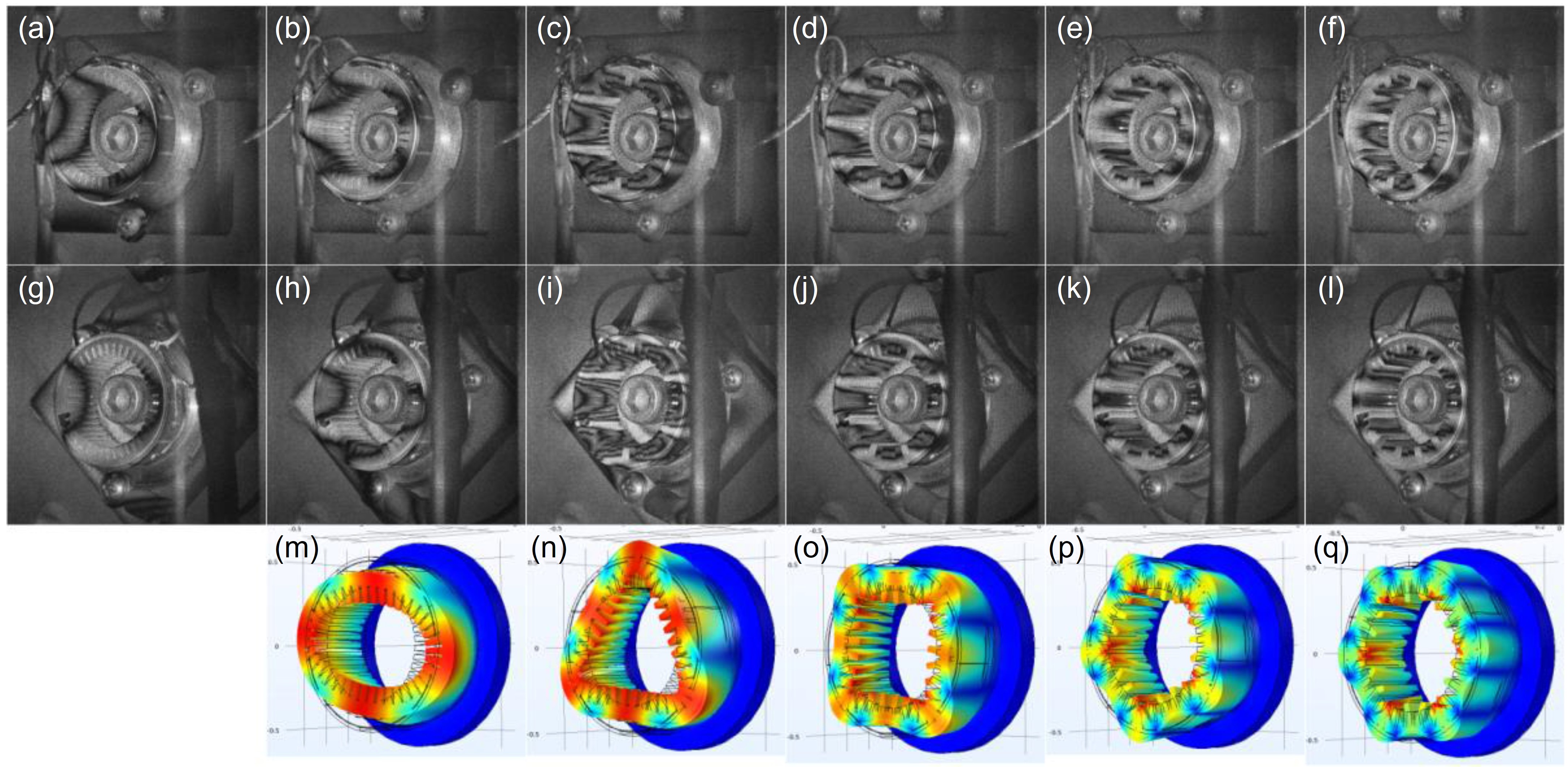}
    \caption{The holography results of both HCMs and compared with simulation pattern. (a) - (f) HCM\#1 from 1$^{st}$ to 6$^{th}$ excitation mode pattern. (g) - (l) HCM\#2 from 1$^{}$ to 6$^{th}$ excitation mode pattern. (m) - (q) FEM eigenfrequency simulation results from 2$^{nd}$ to 6$^{th}$ excitation mode pattern.}
    \label{fig:holography}
\end{figure*}

Desired eigenmodes for the constructed stator were determined by manually sweeping the excitation frequency from 1kHz to 60kHz at 100 Hz increments while imaging in time-averaged mode. The results of deformations in each excitation frequency are shown in Fig \ref{fig:holography}. Unlike the FEM simulation, the 1$^{st}$ eigenmode was able to observe under holographic time-averaged mode. The frequencies between the holography capture and simulation are listed in Table \ref{tab:simvsreal}, and a few kilohertz are acquired between each eigenmode, which is within tolerance. For complete consideration, C360 and pure copper both acquired eigenfrequency results from the same simulation modeling. Both materials are all very close to the holography captured frequencies with the largest difference of 6.1\%. 

\begin{table}[]
\centering
\caption{Time-averaged holography results of HCM 1 \& 2, and eigenfrequency simulation results depicting the deformed stator from 2$^{nd}$ to 6$^{th}$. Unit - kHz}
\label{tab:simvsreal}
\begin{tabular}{c|c|c|c|c}
\hline
Mode\# & HCM\#1 & HCM\#2 & C360 Sim & Copper Sim \\ \hline
1 & 5.910  & 6.040   & - & - \\ \hline
2 & 13.240 & 13.400 & 12.4812  & 12.6032 \\ \hline
3 & 21.880 & 22.860 & 21.8119  & 22.1628  \\ \hline
4 & 31.040 & 32.250 & 30.5002  & 31.0245  \\ \hline
5 & 40.470 & 42.710 & 39.8604  & 40.5951  \\ \hline
6 & 51.930 & 52.310 & 49.1277  & 50.1678  \\ \hline
7 & -      & 65.993 & 63.8972  & 65.5760  \\ \hline
\end{tabular}
\end{table}

\subsection{Rotary Speed Evaluation}

\begin{figure}
    \centerline{\includegraphics[width=.9\linewidth]{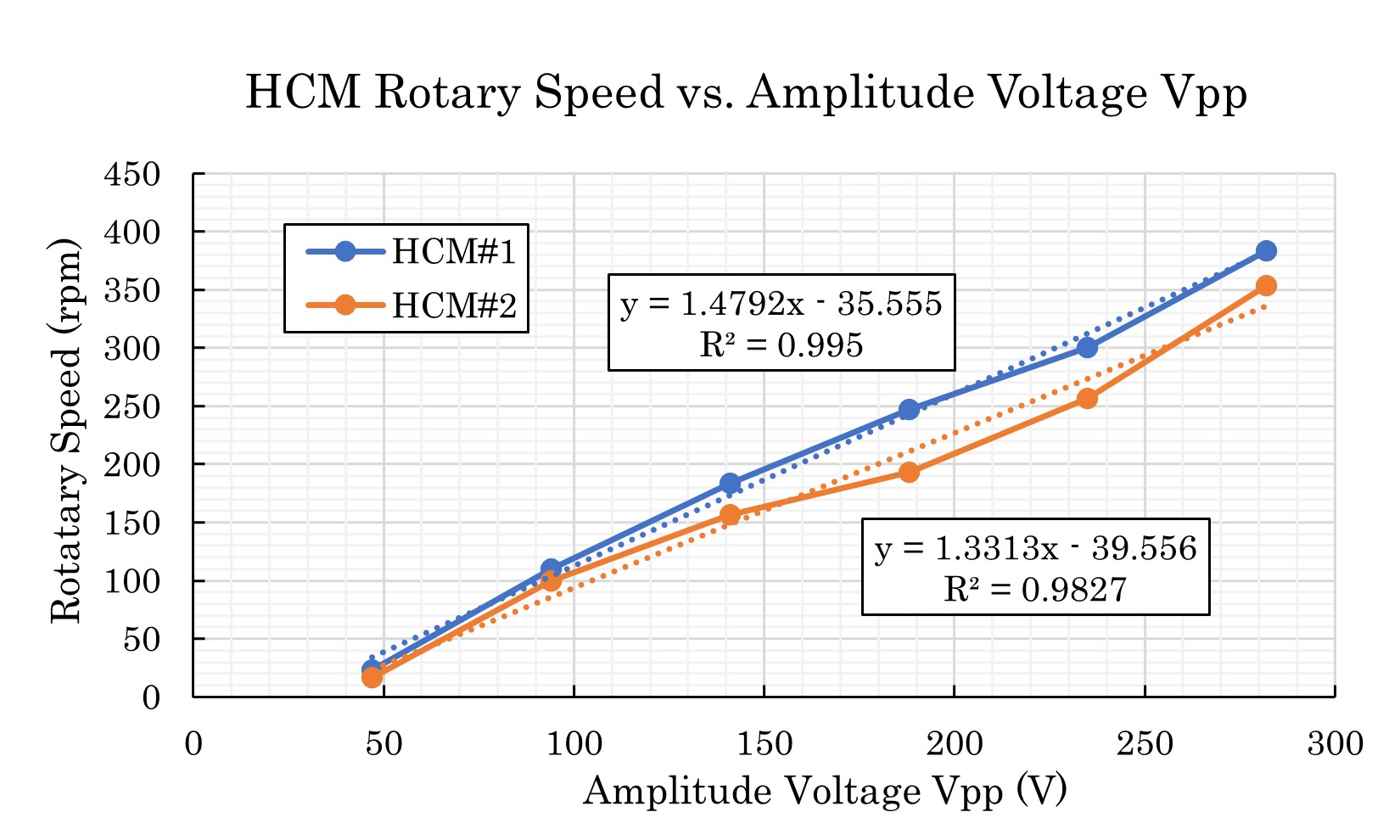}}
    \caption{HCM 1 \& 2 rotary speed measurement under different excitation voltage. The highest rotary speed 383.3333 rpm performed under 282$V_{pp}$ (3$V_{pp}$ on function generator). Both HCMs show a linear relationship between rotary speed and applied voltage, with a coefficient of determination $R^2$ equal to 0.9950 and 0.9827.}
    \label{fig:speed}
\end{figure}

A detailed description of the driving system and performance measurement system can be found in \cite{zhao2024design}. The rotary speed of both HCMs were measured, and the results are shown in Fig \ref{fig:speed}. Both HCMs were tested under zero load and amplified/function generator provided voltage at 46$V_{pp}$/0.5$V_{pp}$, 94$V_{pp}$/1.0$V_{pp}$, 141$V_{pp}$/1.5$V_{pp}$, 188$V_{pp}$/2.0$V_{pp}$, 235$V_{pp}$/2.5$V_{pp}$, and 282$V_{pp}$/ 3.0$V_{pp}$. The highest speeds 383.3333 and 353.3333 rpm with HCM 1 \& 2 respectively performed at 282$V_{pp}$, which is the standard operating voltage amplitude for USR30 series motors. All the rotary speed data was measured at a stable running value. Results suggest that rotary speed is proportional to the applied voltage, with the coefficient of determination $R^2$ equal to 0.9950 and 0.9827 respectively. Based on this result further speed control can be developed.

\subsection{Output Torque Evaluation}

\begin{figure}
    \centerline{\includegraphics[width=.9\linewidth]{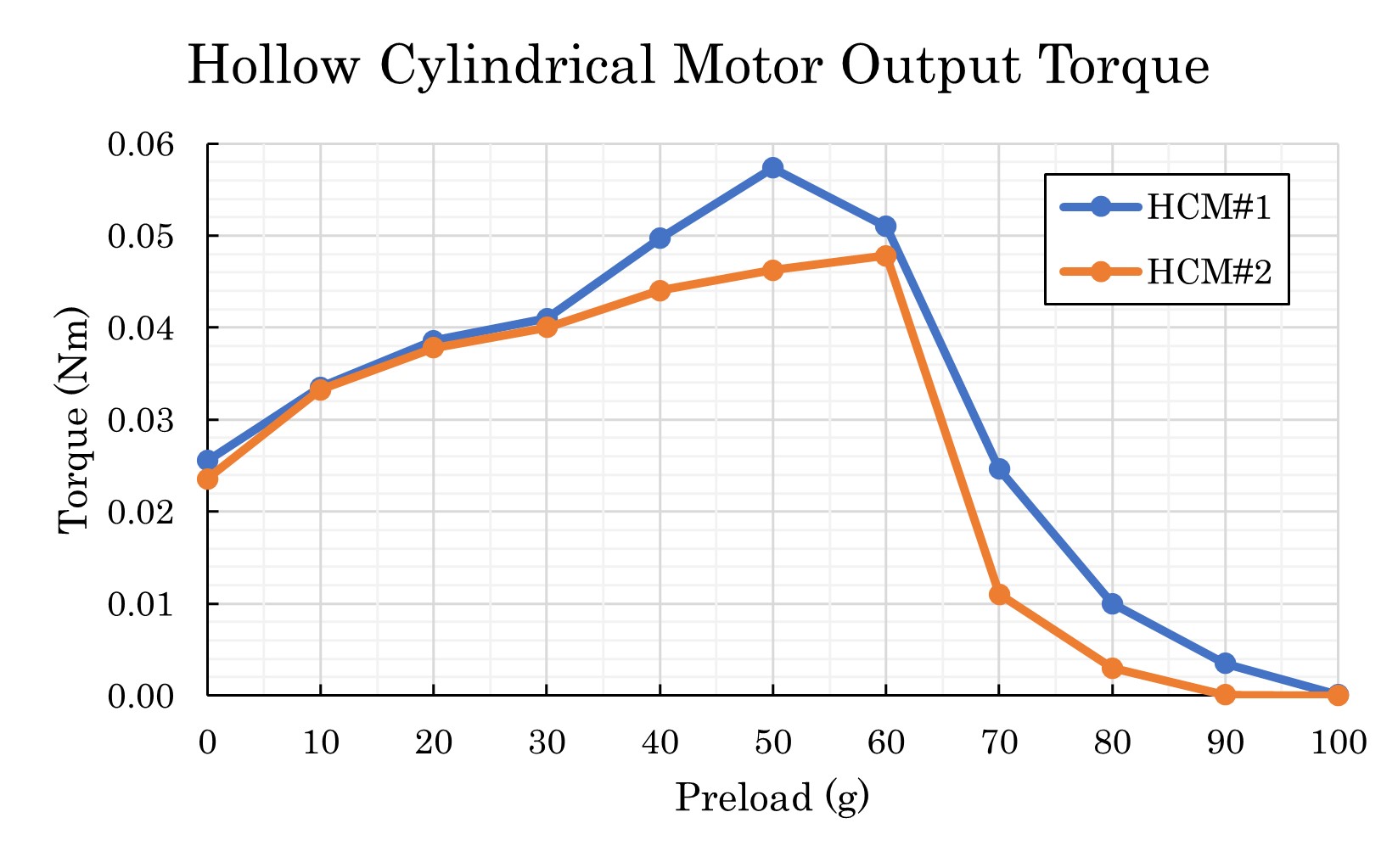}}
    \caption{HCM 1 \& 2 output torque measurement under preload. The Highest torque 57.3504 mNm performed under 50g prepressure with 282$V_{pp}$ driving voltage.}
    \label{fig:torque}
\end{figure}

The measurement of the output torque of HCMs was using the same setup also mentioned in \cite{zhao2024design}. By pulling the cable fixed on the torque-meter shaft, the data collected from a NI USB-6001 Daq card was acquired, where each condition included 5 trials of testing and the results are shown in Fig \ref{fig:torque}. Under zero load, 10g, and 50g preload the motor performed 25.5166/23.5593 mNm, 33.4833/33.2477 mNm, and 57.3504/46.2197 mNm respectively with HCM 1 and 2. Notice that preload over 50g was not measurable because the rotor would not move. A larger output torque may possibly be performed if a larger prepressure is applied.

\section{Conclusion and Disscussion}

In this article, we represented a preliminary custom-made hollow cylindrical motor (HCM) with towards center amplitude and along circumference traveling waves, which was based on USR30 series motors to deal with needle direct rotation motion control. We achieved a design for the new HCM with FEM simulation, holographic interferometry validation, and experimental validation for output performance. Although the DHS was not able to measure the inner surface of the hollow cylindrical design, we still validated the motion with a 45$^\circ$ configuration and collected the desired data. The simulation matched with the holographic images resulting in vibration patterns, and the eigenfrequencies with a largest difference of 6.1\%. The performance of HCMs were measured, which reached up to 383.3333rpm rotary speed under 282V peak-to-peak voltage, and 57.3504mNm torque under 50g prepressure. The rotary speed results were also suggested to be proportional to the voltage applied, which can be used for speed controller design. The highest torque was limited to 50g prepressure, considering the higher prepressure reduced the motion between the stator and rotor. The prepressure was now applied onto the stator with 1$^\circ$ tapered, however, the applied prepressure required a finer solution for better motor performance. Future work can focus on the prepressure applying method, which addresses the limitation of the small prepressure applying issue, and a complete motor design is also under consideration. The friction layer is not discussed in this work, the utilization of the friction layer, or making this motor Ultem plastic to address more feasibility of MRI compatibility should be further studied.

\bibliographystyle{IEEEtran}
\bibliography{Main}

\end{document}